\documentclass[11pt]{article}
\RequirePackage[table]{xcolor}
\RequirePackage{booktabs}
\RequirePackage{multirow}
\RequirePackage{caption}% typesetting of captions
\usepackage{amsmath, amsthm, amssymb}
\usepackage{graphicx,multirow}
\usepackage{verbatim}
\usepackage{natbib}
\usepackage{caption}
\usepackage{subcaption}
\usepackage{fancyvrb}
\usepackage{enumerate}
\usepackage{relsize}
\usepackage{float}
\usepackage{setspace}
\usepackage[utf8]{inputenc}
\usepackage{algorithm}
\usepackage{algpseudocode}
\usepackage{xcolor}
\usepackage{tikz}
\usepackage{multirow}
\usepackage{hyperref}
\usepackage{setspace}
\usepackage[margin=1in]{geometry}
\hypersetup{colorlinks,citecolor=blue,urlcolor=blue,linkcolor=blue}

\usepackage{max_tex}
\usepackage{hyperref}
\usepackage{geometry}
\usepackage{pdflscape}
\graphicspath{{./figures/}}

\theoremstyle{plain}

\theoremstyle{definition}

\newtheorem{defi}{Definition}
\newtheorem{assu}{Assumption}
\newtheorem{rema}{Remark}

\newfloat{algbox}{tbp}{lop}
\newtheorem{alg}{Algorithm}

\author{Maximilian Schuessler\thanks{Department of Biomedical Data Science, Stanford University, Stanford, CA; 
ORCID ID: 0000-0002-8641-783X}  
\\\texttt{maxsc@stanford.edu}
\and
Erik Sverdrup\thanks{Department of Econometrics \& Business Statistics, Monash University, Clayton, VIC; 
ORCID ID: 0000-0001-6093-1390}  
\\\texttt{erik.sverdrup@monash.edu}
\and
Robert Tibshirani\thanks{Departments of Biomedical Data Science and of Statistics, Stanford University, Stanford, CA;
ORCID ID: 0000-0003-0553-5090} 
\\ \texttt{tibs@stanford.edu}}

\date{\ifcase\month\or
January\or February\or March\or April\or May\or June\or
July\or August\or September\or October\or November\or December\fi \ \number%
\year}

\title{Statistical Learning for Heterogeneous Treatment Effects: Pretraining, Prognosis, and Prediction}

\begin{document}

\maketitle
\begin{abstract}
\noindent Robust estimation of heterogeneous treatment effects is a fundamental challenge for optimal decision-making in domains ranging from personalized medicine to educational policy. In recent years, predictive machine learning has emerged as a valuable toolbox for causal estimation, enabling more flexible effect estimation. However, accurately estimating conditional average treatment effects (CATE) remains a major challenge, particularly in the presence of many covariates. In this article, we propose pretraining strategies that leverage a phenomenon in real-world applications: factors that are \emph{prognostic} of the outcome are frequently also \emph{predictive} of treatment effect heterogeneity. In medicine, for example, components of the same biological signaling pathways frequently influence both baseline risk and treatment response. Specifically, we demonstrate our approach within the  $R$-learner framework, which estimates the CATE by solving individual prediction problems based on a residualized loss. We use this structure to incorporate side information and develop models that can exploit synergies between risk prediction and causal effect estimation. In settings where these synergies are present, this cross-task learning enables more accurate signal detection, yields lower estimation error, reduced false discovery rates, and higher power for detecting heterogeneity.
\end{abstract}

\noindent \textbf{Keywords}: Statistical learning, causal inference, heterogeneous treatment effects, biomarkers \\

% \hfill word count: 5,347  % Remove for arXiv

\section{Introduction}
\label{sec:Introduction}
For many applications, from personalized medicine to educational policies, understanding treatment effects beyond global averages is critical for optimal decision-making and treatment prioritization. For example, a cancer treatment may show average benefits for a patient population, yet individual responses can vary widely among patients \citep{tomasik2024heterogeneity}. In the presence of such heterogeneity, the average may be composed of individuals with large, moderate, and negative treatment effects \citep{kravitz2004evidence}. Those with minimal benefit or harm are better off receiving no or alternative treatments. This explains why in medicine and other high-stakes settings, relying solely on average treatment effects (ATE) can lead to suboptimal decisions--potentially harming some individuals or, incorrectly, withholding treatment from those who could benefit. To advance precision medicine, researchers and practitioners are, therefore, interested in developing as highly personalized treatment recommendations as possible \citep{kessler2021pragmatic, kosorok2019precision}. One common strategy for deriving personalized treatment policies is to use a risk-based approach, where individuals prone to poor outcomes receive treatment first \citep{kent2016risk, kent2020predictive}. An alternative approach is to prioritize individuals based on their treatment benefit \citep{manski2004statistical, wager2018estimation}. 

In settings with expected differential treatment responses, knowing each unit's individual treatment effect (ITE) enables the design of personalized treatment policies, prioritization rules, and helps prevent under- or over-treatment. In practice, however, the ITE is not point-identified \citep{imbens2015causal}. A common strategy to obtain more granular treatment effects is to estimate the conditional average treatment effect (CATE), which is still an average treatment effect but conditional on a set of covariates. Conditioning on more covariates makes this average more specific, potentially revealing subgroups with treatment effects that differ significantly from the overall average.

With the widespread availability of large-scale observational data, researchers now have access to rich data environments that offer the potential to estimate the conditional average treatment effects for increasingly granular subgroups. Simultaneously, machine learning has emerged as a useful toolbox, offering several advantages for CATE estimation \citep{athey2016recursive}. A large body of work has made considerable progress towards more flexible estimation, e.g., by using the lasso \citep{RN127}, boosting \citep{powers2018some}, multivariate adaptive regression \citep{powers2018some}, neural networks \citep{shalit2017estimating}, and random forests \citep{wager2018estimation}, to name just a handful of approaches. Moreover, these models can be flexibly integrated into metalearners \citep{kennedy2023towards, kunzel2019metalearners, morzywolek2023general, nie2021quasi, van2024combining} that decompose the exercise of effect estimation into separate prediction tasks. This has enhanced the role of statistical learning for CATE estimation in recent years.

Beyond more flexible estimation, automated CATE models offer a principled and objective way of identifying and reporting subgroups that drive heterogeneous treatment responses \citep{athey2019generalized}. This provides a remedy to more traditional approaches in which practitioners need to define relevant subgroups \textit{a priori}. Moreover, it mitigates practices similar to $p$-hacking, where possible subgroups are repeatedly searched until a putative signal for heterogeneity is found. Especially in areas like genetics, where treatment heterogeneity is often the result of complex interactions, such manual approaches are not scalable and become easily impractical.

Despite these advances, there remains a fundamental statistical difficulty in CATE estimation \citep{kennedy2024minimax}. Identifying heterogeneous treatment effects is especially challenging when sample sizes are limited (e.g., in rare diseases) or when effects are diffuse and involve complex interactions. Identifying effect modifiers then often resembles the task of finding a ``needle in a haystack''. For example, in settings with $p = 10$ binary covariates, there are already $2^{10} = 1024$ possible subgroups that could explain heterogeneity. The challenge grows in high-dimensional settings like gene expression data, where $p$ can be as large as, or even exceed $n$. In those settings, low power, overfitting, and high variance are concerns that complicate estimating the CATE \citep{hastie2009elements, powers2018some, candes2018panning}.

Low sample sizes and high-dimensional data are common challenges in medical settings. For instance, clinical trials increasingly include more covariates (e.g., from molecular data), while their sample sizes have not kept pace. As a result, there is an urgent need for methods that improve upon traditional approaches to CATE estimation. \citet{ignatiadis2021covariate} suggests mitigating these challenges by leveraging domain knowledge and hidden side information in the data---for example, to derive data-driven weights for hypothesis testing. In a similar vein, \citet{fortney2015genome} and \citet{dobriban2015optimal} show how prior information on disease-related gene loci may help elucidate loci for longevity due to their existing overlap. Seemingly unrelated disease loci can predict treatment responses across multiple conditions. These examples highlight that real-world datasets in biomedicine, though often characterized by a low signal-to-noise ratio, are rich in connections and structures. These might involve risk measures that serve as surrogates for treatment effects and help uncover effect modifiers. In predictive settings, statistical learning methods that incorporate weights from related tasks—commonly known as transfer learning—have gained significant attention over the past decades \citep{argyriou2006multi,  craig2024pretraining, gu2024robust, obozinski2010joint, zou2006adaptive,li2022transfer}. Harnessing putative side information for heterogeneous treatment effects bears considerable potential, but requires statistical learning strategies that can synergize seemingly distinct prediction tasks for CATE estimation. 

In this article, we use the term ``pretraining'' as an umbrella term to denote various statistical learning procedures that enable a sequential transfer of information from one (or multiple) prediction model(s) to another. Depending on the architecture of each model, this can take the form of using the active set from a lasso to derive penalty factors for the second model (adaptive lasso), performing weighted feature sampling based on feature importance in tree-based learners, or transferring weights to a neural network.

We leverage recent advances in statistical learning for formulating the CATE estimation as a causal loss minimization objective. We focus our exposition on the $R$-learner and exploit a common phenomenon in real-world settings: factors that are prognostic of the mean outcome are frequently predictive of the treatment effect. We claim that in settings in which covariates have this dual role of being prognostic \emph{and} predictive, we can improve CATE estimation by synergizing predictive learning tasks. We refer to this approach as ``pretraining the $R$-learner''. In addition, we showcase how this approach can be generalized for various linear and non-linear learners and their statistical learning properties. In a nutshell:

\begin{enumerate}
    \item[(a)] We propose a pretraining strategy that goes beyond treating the mean outcome function as a mere nuisance parameter in the $R$-learner framework: we leverage the shared support between prognostic and predictive factors for conditional average treatment effect estimation.
    \item[(b)] The goal of this approach is threefold: first, increase the accuracy of CATE estimation by exploiting synergies between seemingly independent prediction tasks; second, improve the support recovery of effect modifiers/interacting effects; and third, gain more insights about the putative existence of shared support between prognostic and predictive factors when estimating the CATE. 
    \item[(c)] To demonstrate this approach, we develop a suite of estimation frameworks using the lasso-based $R$-learner ($R$-lasso) and non-parametric models. We also demonstrate how this approach can be extended to non-linear settings using basis function expansions and random forests.

\end{enumerate}

\section{Leveraging Side Information for Heterogeneous Treatment Effect Estimation} 
Let $\{Y_i(0), Y_i(1)\}$ denote potential outcomes in the control and treatment state \citep{imbens2015causal}. Suppose $n$ units of observation resulting in tuples $\{Y_i, W_i, X_i\}_{i=1}^n$, where $Y_i = Y_i(W_i)$ corresponds the $i$-th unit's realized outcome, $W_i = \{0,1\}$ denotes the treatment assignment, and $X_i\in \mathcal{X}$ are individual-level features, where $\mathcal{X}$ is some potentially large $p$-dimensional feature space. We make the following assumptions.

\begin{assu} \label{asu:ignorable}
    (Unconfoundedness). Conditional on a set of observed covariates $X_i$, treatment assignment $W_i$ is independent of $Y_i$,
    $$
    \{Y_i(0), Y_i(1)\} \indep W_i \mid X_i.
    $$
\end{assu}

\begin{assu} \label{asu:positivity}
    (Positivity). The propensity score $e(x) = \PP{W_i = 1 \mid X_i = x}$, which defines the probability of exposure to treatment conditional on covariates, is bounded away from 0 and 1. There exists an $\eta > 0$ such that
    $$
    \eta \leq e(x) \leq 1-\eta ~\text{for all}~ x \in \mathcal{X}.
    $$
\end{assu}

\begin{rema}
    To simplify notation, we use the symbol $X_i$ to refer to both potential confounders, treatment effect modifiers, etc. We emphasize this distinction when it matters.
\end{rema}

The ideal clinical decision making quantity is the $i$-th unit's treatment effect, $Y_i(1) - Y_i(0)$. By the fundamental problem of causal inference, this quantity, however, is not point-identified because each unit $i$ can either receive treatment or not. In order to gain insight into personalized treatment responses, it is helpful to instead shift attention to conditional average treatment effects (CATE), defined as $\tau(x) = \EE{Y_i(1)- Y_i(0) \mid X_i = x}$. We denote the mean conditional potential outcome in arm $w$ by $\mu_{(w)}(x) = \EE{Y_i(w) \mid X_i = x}$. Given Assumption \ref{asu:ignorable}-\ref{asu:positivity}, $\mu_{(w)}(x) = \EE{Y_i \mid X_i=x, W_i = w}$, and the CATE is identified by
\begin{equation}
\tau(x) = \EE{Y_i \mid X_i=x, W_i = 1} - \EE{Y_i \mid X_i=x, W_i = 0}.
%\tau(x) = \mu_{(1)}(x) - \mu_{(0)}(x),
\end{equation}
An immediate strategy to obtain estimates of CATE via this identity is to estimate $\mu_{(0)}(X_i)$ and $\mu_{(1)}(X_i)$ via predictive machine learning algorithms fitted separately on each treatment arm, followed by estimating the CATEs through their plug-in difference $\hat \tau(X_i) = \hat \mu_{(1)}(X_i) - \hat \mu_{(0)}(X_i)$. This approach is usually referred to as the $T$-learner \citep{kunzel2019metalearners}.

A drawback of this simple approach is regularization bias due to, for example, treatment imbalance \citep{kunzel2019metalearners}. Recent advances have produced an array of improvements to this approach by decomposing prediction tasks into tailored loss functions that directly target the CATE (e.g., \citet{foster2023orthogonal, kennedy2023towards, kunzel2019metalearners, morzywolek2023general, nie2021quasi, van2024combining}). The goal of these approaches is to strip out confounding and baseline effects and then focus on the CATE target parameter. Common in these loss-based approaches are nuisance components that we might leverage for the final CATE goal. In particular, the $R$-learner \citep{nie2021quasi} leverages Robinson's transformation \citep{robins2004optimal, robinson1988root} to express the CATE function as
\begin{equation}
    \tau(\cdot) = \argmin_{\tau} \bigg\{ \EE{\bigg[(Y_i - m(X_i)) - (W_i - e(X_i)) \tau(X_i)\bigg]^2} \bigg\}, 
\end{equation}
where $m(x) = \EE{Y_i \mid X_i=x}$ is the mean conditional outcomes marginalized over treatment. To estimate CATE via this approach, we form suitable cross-fit/sample-split estimates $\hat m(X_i)$ and $\hat e(X_i)$ of the mean outcomes and propensity scores, then plug them into the corresponding empirical loss\footnote{Whenever an estimated quantity, such as $\hat m(X_i)$, enters the right-hand side of an expression, we abstain from resorting to additional notation and maintain the convention that the $i$-th predictions are constructed without using the $i$-th unit for estimation, such as with, for example, cross-fold estimation.}, which we minimize over some chosen CATE function class:
\begin{equation}\label{eq:empirical_rmin}
    \hat \tau(\cdot) = \argmin_{\tau} \Bigg\{ \frac{1}{n}\sum_{i=1}^{n}\bigg[(Y_i - \hat m(X_i)) - (W_i - \hat e(X_i)) \tau(X_i)\bigg]^2  \Bigg\}.
\end{equation}
This loss function satisfies an oracle property that roughly guarantees that as long as $\hat m(X_i)$ and $\hat e(X_i)$ are estimated reasonably well, we get similar performance to as if we had access to true $m(X_i)$ and $e(X_i)$. This empirical $R$-loss yields a tunable causal loss,
\begin{equation} \label{eq:rloss}
    \widehat L_n = \frac{1}{n}\sum_{i=1}^{n}\bigg[(Y_i - \hat m(X_i)) - (W_i - \hat e(X_i)) \hat \tau(X_i)\bigg]^2,
\end{equation}
which allows us to tune hyperparameters in the learner for $\tau(x)$. If certain features in the support of $m(X_i)$ are also predictive of $\tau(X_i)$, then 
the loss of this CATE function should ideally reflect the benefits from leveraging prognostic side information in a data-driven way. 

Another way of expressing the CATE via a data-dependent loss with similar oracle guarantees is the $DR$-learner \citep{kennedy2023towards} that leverages \citet{robins1994estimation}'s augmented inverse-propensity weighted scores to fit the CATE via the following pseudo-outcome regression
\begin{equation}\label{eq:drlearner}
{\hat \tau(\cdot) = \argmin_{\tau} \Bigg\{\sum_{i = 1}^{n} \bigg[
\left(
\hat{\mu}_{(0)}(X_i) + \hat{\mu}_{(1)}(X_i) +
\frac{W_i - \hat{e}(X_i)}{\hat{e}(X_i)(1-\hat{e}(X_i))}(Y_i-\hat{\mu}_{(W_i)}(X_i))
\right) - \tau(X_i)
\bigg]^2}\Bigg\}.
\end{equation}
One particularly useful ingredient in this loss is the model for the conditional mean in the absence of treatment, $\hat{\mu}_{(0)}(X_i)$. If features that are predictive of baseline outcomes also matter for treatment benefits, then sharing information between $\hat{\mu}_{(0)}(X_i)$ and $\hat \tau(X_i)$ might be advantageous in estimating the CATE. One downside of this approach is that it effectively splits the pretraining exercise by treatment assignment, reducing the amount of data available for pretraining from each nuisance parameter. Moreover, pretraining solely on the samples assigned to the control risks overlooking treatment-induced signal in $Y_i$, thereby sacrificing a potential safeguard against harmful pretraining, which is automatically built in when pretraining on the mean outcome function in the $R$-Learner. 

For these reasons and brevity, we restrict our study to exploring leveraging the $R$-learner to share information between marginalized outcomes $m(X_i)$ and the CATE. In the supplementary information, we showcase that our logic for pretraining also holds for the $DR$-Learner. For a general discussion and comparison of orthogonal losses for CATE estimation, see, for example, \citet{fisher2024inverse, morzywolek2023general}.

\subsection{Shared Support between Prognostic and Predictive Factors} \label{section:support}
There are many ways in which information can be shared between two statistical tasks to yield a better outcome. In this study, we explore mechanisms that leverage the ``shared support'' of two prediction tasks. We define shared support as a setting in which two or more prediction tasks operate over a common active feature space--that is, a subset of input variables that are informative and utilized across tasks. While this requires overlap in the predictors, the targets may differ.

A common observation in medicine and other fields is that factors that are prognostic of an outcome are also predictive of treatment response, i.e. there is a overlap in the support of features that are relevant to solve the objective function for the mean outcome function $m(X_i)$ and those for the CATE function $\tau(X_i)$. Equation \ref{eq:mean_outcome} expresses this phenomenon in an easy toy example based on a linear model. The covariate $X_{i1}$ is simply \emph{prognostic} of the outcome $Y_i$  (in the absence of the treatment); $X_{i3}$ is \emph{predictive} as it interacts with the treatment. Finally, $X_{i2}$ is both prognostic and predictive.
\begin{equation}
    Y_i = \beta_0 + X_{i1} \beta_1 + X_{i2} \beta_2 + W_i \tau(X_i) + \varepsilon_i \\
    \label{eq:mean_outcome}
\end{equation}
\begin{equation}
    \tau(X_i) = \theta_0 + X_{i2} \theta_2 +  X_{i3} \theta_3
    \label{eq:tau_funcrtion}
\end{equation}
\begin{defi} \label{defi:Prognostic}
    (Prognostic Factor). A prognostic factor is any covariate or combination of covariates that contains information about the likelihood of an outcome, independent of any treatment assignment \citep{ballman2015biomarker, nci_prognostic_factor}.
\end{defi}

\begin{defi} \label{defi:Predictive}
    (Predictive Factor). A predictive factor is any covariate or combination of covariates that contains information about the likelihood of the treatment effect $\tau(X_i)$. A predictive factor interacts with the treatment. This interaction can be \emph{quantitative}, where both groups (with and without the prognostic factor) benefit from the treatment but at varying degree; or \emph{qualitative}, that is either the group with or without the predictive factor derives a positive effect from the treatment, but not both \citep{ballman2015biomarker, nci_predictive_factor}.
    
\end{defi}

\begin{figure}[H]
  \centering
  \makebox[\textwidth][c]{\includegraphics[width=1.15\textwidth]{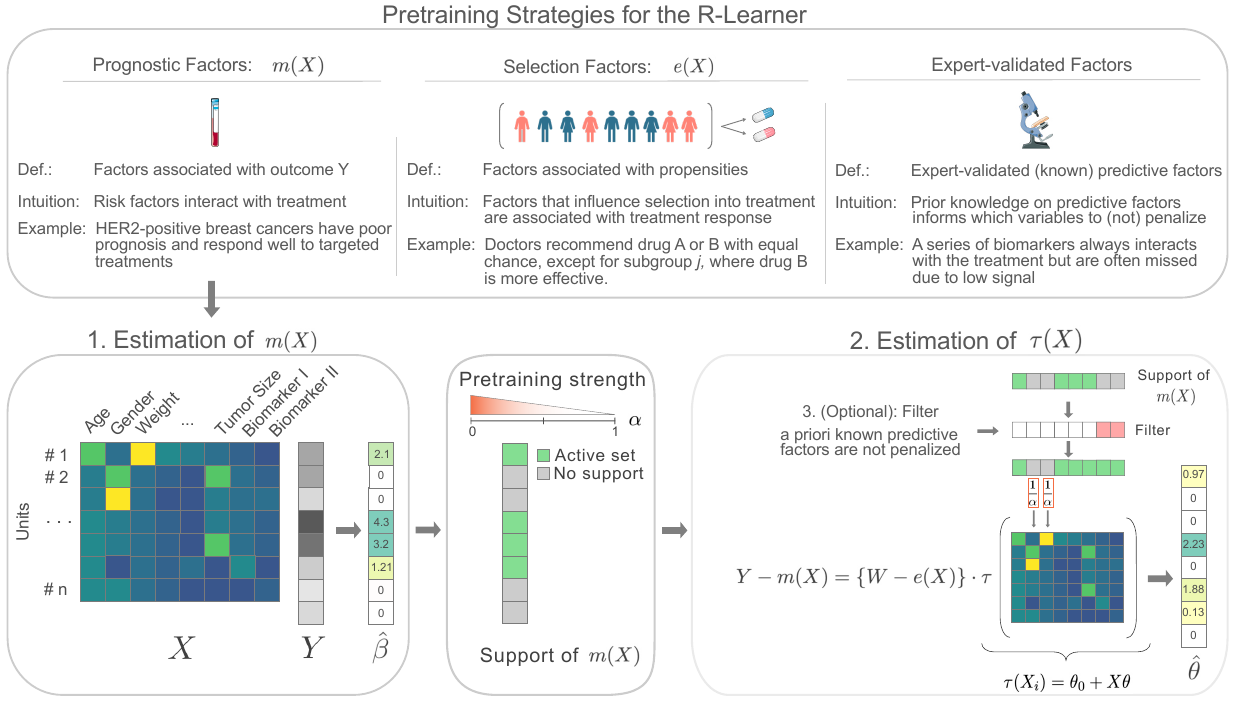}}
  \caption{\textbf{Overview of the Pretrained $R$-learner.} The figure shows the step-wise procedure of the $R$-learner with the mean outcome function and the CATE. The upper panel shows different pretraining strategies. The lower panels illustrate how these pretraining approaches are integrated into the $R$-learner using the mean outcome function as an example. The active set for the mean outcome model (left panel) is used for pretraining to inform the estimation of the conditional average treatment effect (CATE) model. Here, gender, weight, and biomarker II are not in the support of the mean outcome model (middle panel) and get penalized at $1/\alpha$ when estimating the CATE model (right panel). Optionally, previously known, expert-validated predictive factors are exempt from any penalties. For example, biomarker II is known to be predictive (and thus protected from any penalties a priori), but not in the active set of the mean outcome model. Age, tumor size, and biomarker I are in the support of both models and do not get penalized through pretraining. Icons from Servier Medical Art, licensed under CC BY 4.0,  https://creativecommons.org/licenses/by/4.0/ .} 
  \label{fig:pretraining_illustration} 
\end{figure}
A classic example in medicine for the presence of this joint support is Human Epidermal Growth Factor Receptor 2 (HER2): individuals with a breast cancer of this type, which occurs in approximately 20\% of all cases, have a higher risk of relapse and shorter overall survival \citep{slamon1987human}. HER2 expression is associated with increased resistance to chemotherapy \citep{slamon2001use}, while enhancing sensitivity to specific subclasses of therapies targeted against HER2 \citep{hudis2007trastuzumab}. Thus, HER2 is both a prognostic factor for survival outcomes $Y_i$ and predictive of the treatment effect $\tau(X_i)$. In addition to single biomarkers, there is an increasing number of composite biomarkers such as gene signatures that determine risk and treatment response in oncology \citep{chapuy2018molecular, alizadeh2000distinct, rueda2019dynamics}.

The overlap of predictive and prognostic factors also plays a role in psychiatry and cardiology: patients with depression have a higher risk and poorer outcomes for cardiovascular diseases (CVD) \citep{lichtman2014depression}. Furthermore, patients who suffer from depression are at higher risk for medication non-adherence, which can decrease the effectiveness of CVD medications such as statins and anti-hypertensive treatments \citep{dimatteo2000depression, carney2017depression}.
Finally, the phenomenon of shared support is likely to extend to many other disciplines outside medicine. \citet{athey2025machine} demonstrates that baseline responses are easier to estimate than treatment effects and leverages this for enhanced causal predictions to target behavioral nudges. In education, for example, school-specific characteristics, such as learning environments, not only predict students' overall educational performance, but also often modify the effects of educational interventions \citep{yeager2019national}.

\subsection{Side Information from Selection Factors}
It is possible to extend our concept of pretraining to the propensity score function in observational settings where treatment assignment is the function of some covariates. Suppose clinicians tend to recommend a treatment to patients in an age group where they believe it is most effective. These patients will have high treatment propensity. If clinicians are generally correct, then the same covariates driving propensity will also predict treatment benefit. \citet{cohen2018treatment} discusses the role of prognostic, predictive, and selection factors in the context of depression. Selection factors also play an important role in phase-II and phase-III cancer trials, where investigators choose one (``physician's choice'') among many treatment options in the control arm \citep{modi2022trastuzumab, ribas2015pembrolizumab}. 
  \begin{defi} \label{defi:selection}
    (Selection Factor). A \emph{selection factor} is any pre-exposure covariate that contains information about the likelihood of treatment assignment. 
\end{defi}

Pretraining on selection factors might also be relevant in other areas for research, such as education. There is a large body of research examining the effect of learning interventions or college admission on outcomes (e.g., future income). In this example, selection factors such as socio-economic status, motivation, beliefs about one's future, and health might be strong effect modifiers \citep{yeager2016using}.

\section{Using the Lasso to Leverage Prognostic Side Information}\label{section:rlassolinear}
\begin{figure}[H]
  \centering
  \makebox[\textwidth][c]{\includegraphics[width=1.0\textwidth]{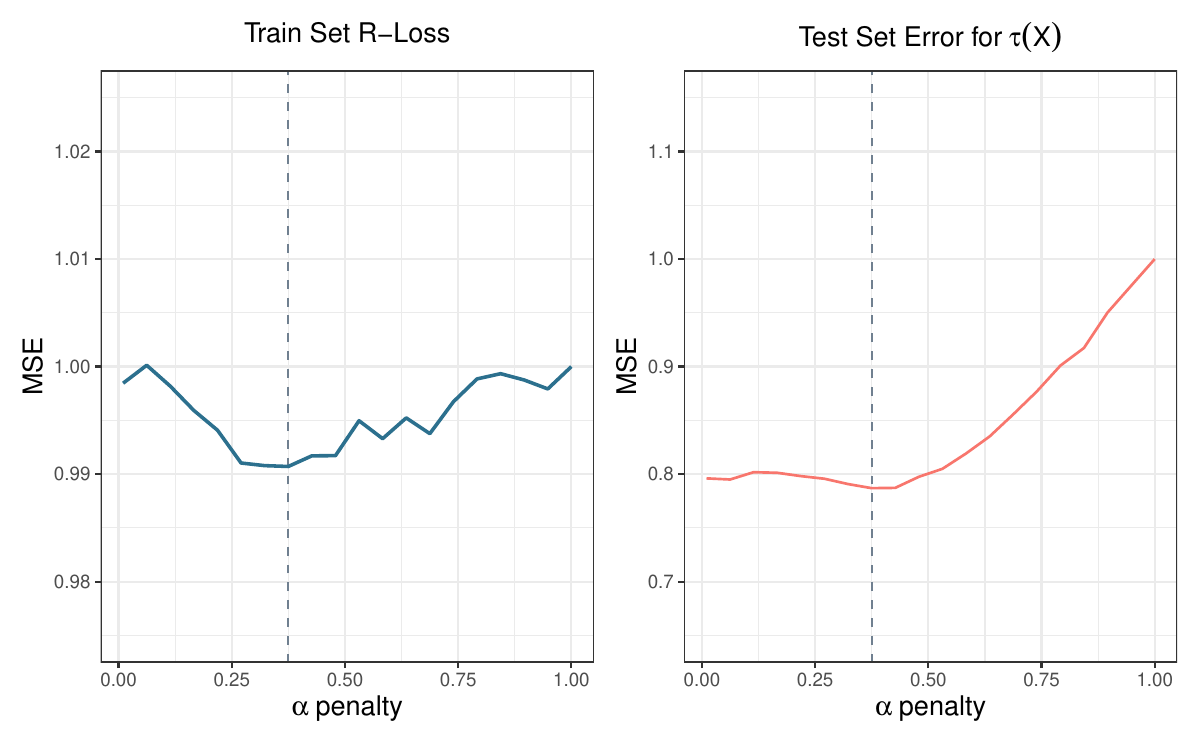}}
  \caption{\textbf{Hyperparameter Tuning Using the R-Loss.} The figure shows the  median performance of the $R$-lasso in terms of $R$-loss on the training set and MSE on the test set in a linear setting with $n = 500$, $p = 150$, $SNR = 2$, an active set of size $|S| = \frac{p}{3}$ and for $5000$ repetitions. The dashed vertical lines show the $\alpha$s for which the train set $R$-loss and test set MSE are minimized.}
  \label{fig:fig_rloss_illustration}
\end{figure}

In order to instantiate an $R$-learner that leverages shared support, we need learners for the mean conditional outcome function and CATE function amenable to this support-sharing task. One tractable solution is to leverage an adaptive lasso for the CATE function, where the differential shrinkage parameters are set through the lasso active set solution for the mean conditional outcome function. 
For exposition, it is helpful to first consider regularized high-dimensional linear models. Suitable featurizations of the covariate matrix \citep{hastie2009elements} allow us to extend this approach to non-linear settings as revisited in \autoref{section:non_linear}. We start out by assuming a randomized controlled trial setting where the propensity scores are known and omit this estimation task. We start by fitting $m(X_i)$ with the lasso
\begin{align} \label{eq:mlasso}
% \begin{split}  
m(X_i) = \beta_0 + \sum_{j=1}^p X_{ij}\beta_j, \\
\hat{\beta} = {\argmin}_{\beta} \Bigg\{  \sum_{i=1}^n \Big[Y_i- m(X_i)\Big]^2 + \lambda_1 \sum_{j=1}^p |\beta_j | \Bigg\}, \nonumber
% \end{split}
\end{align}
where $\lambda_1$ is a regularization parameter that minimizes the cross-validated MSE. In addition to the estimated model coefficients $\hat \beta$, the lasso yields an active set of features with nonzero coefficients. If some of these coefficients are not part of the active set, then, as argued in Section \ref{section:support}, they might be less useful for estimating the CATE. In the next step, we incorporate information from this active set by estimating $\tau(X_i)$ via an adaptive lasso using \eqref{eq:empirical_rmin}
\begin{align} \label{eq:ptrlasso}
\tau(X_i) &= \theta_0 + \sum_{j=1}^{p} X_{ij}\theta_j, \\
\hat{\theta} = \argmin_{\theta} \Bigg\{ \frac{1}{n} \sum_{i = 1}^{n} \Big[ (Y_i - \hat m(X_i)) &- (W_i - \hat e(X_i)) \tau(X_i) \Big]^2 + \lambda_2 \sum_{j=1}^{p} w_j | \theta_j | \Bigg\}. \nonumber
\end{align}
Here, the regularization parameter $\lambda_2$ applies equally to each feature, but the weights $w_j$ control the relative regularization strength depending on the estimated conditional mean model \eqref{eq:mlasso} support,
\begin{equation} \label{eq:msupport}
w_j =
\begin{cases}
1/\alpha, & \text{if } \hat{\beta}_j = 0 \\
1, & \text{if } \hat{\beta}_j \neq 0.
\end{cases}
\end{equation}
The additional tuning parameter $\alpha$ controls how much differential regularization to place on factors that are not estimated to be prognostic, and we refer to this as the ``pretraining strength'' \citep{craig2024pretraining}. As we lower $\alpha$ from 1 (no penalty) to 0 (penalty towards $\infty$), the strength of the penalty factor increases until the feature is effectively removed from the model. Using the $R$-loss allows us to treat $\alpha$ and thereby the pretraining strength as a tunable hyperparameter depending on the size of the overlap between $m(X_i)$ and $\tau(X_i)$. Figure \autoref{fig:fig_rloss_illustration} shows the $R$-loss on the training set and mean squared error (MSE) on the test set for varying $\alpha$ in a simulation setting with partially shared support. The $R$-loss serves as a reliable loss for tuning, as both errors reach their minimum at $\alpha = 0.375$.

In empirical settings in which the assumption that high risk (prognosis) entails high treatment benefit proves wrong (for empirical examples see \citet{earle2008aggressiveness} or \citet{weeks1998relationship}), we expect the $R$-loss to favor $\alpha$'s closer to 1, removing the pretraining effect. Algorithm \ref{alg:1} summarizes the pretrained $R$-lasso.

\begin{rema}
We can employ a similar strategy in the $DR$-learner \eqref{eq:drlearner} using its nuisance parameters for pretraining. While pretraining based on the conditional outcome model for the control arm $\mu_{(0)}(X_i)$ is appealing, it has one disadvantage: depending on the fraction of treated units, we will have less data at our disposal. Moreover, pretraining using $m(X_i)$ has the potential to pick up signal that stems from $\tau(X_i)$ given that under Assumption \ref{asu:ignorable} we have that $m(X_i) = \mu_{(0)}(X_i) + e(X_i) \tau(X_i)$. This provides a potential safeguard against penalizing predictive factors that are not prognostic when there is little to no overlap between the two types of covariates.
\end{rema}

\begin{algorithm}[H]
\caption{\textbf{: Pretraining the $R$-learner Using the Lasso}}
\begin{algorithmic} % The number tells where the line numbering should start
\State 
\State \textbf{Input: } Data $\{Y_i$, $X_i$, $W_i\}_{i=1}^{n}$, grid of tuning parameters \{$\alpha \in (0, 1]$,~ $\lambda_2$\}.\\

\State \textbf{1.} Construct cross-fit estimates of the mean outcome function $m(X_i)$ with the lasso.
\begin{align*}
% \begin{split}  
m(X_i) = \beta_0 + \sum_{j=1}^p X_{ij}\beta_j, \\
\hat{\beta} = {\argmin}_{\beta} \Bigg\{  \sum_{i=1}^n \Big[Y_i- m(X_i)\Big]^2 + \lambda_1 \sum_{j=1}^p |\beta_j | \Bigg\}.
% \end{split}
\end{align*}

\State \textbf{2.} Construct feature-specific penalty factors $w_j$ using estimates $\hat \beta$ and $\alpha$.
\begin{equation*} 
w_j =
\begin{cases}
1/\alpha, & \text{if } \hat{\beta}_j = 0 \\
1, & \text{if } \hat{\beta}_j \neq 0.
\end{cases}
\end{equation*}\\

\State \textbf{3.} Set the propensity scores $e(X_i)$ 
\smallskip
\If{unknown}
\State  Construct cross-fit estimates of $e(X_i)$ using a logistic lasso.
\Else
\State Set $\hat e(X_i)$ to the known randomization probabilities.
\EndIf 
\\

\State \textbf{4.} Fit $\tau(X_i)$ using $w_j$-weighted lasso with common regularization $\lambda_2$.
\begin{align*} \label{eq:ptrlasso}
\tau(X_i) &= \theta_0 + \sum_{j=1}^{p} X_{ij}\theta_j, \\
\hat{\theta} = \argmin_{\theta} \Bigg\{ \frac{1}{n} \sum_{i = 1}^{n} \Big[ (Y_i - \hat m(X_i)) &- (W_i - \hat e(X_i)) \tau(X_i) \Big]^2 + \lambda_2 \sum_{j=1}^{p} w_j | \theta_j | \Bigg\}.
\end{align*}

\State \textbf{Return:} the estimated $\hat \tau(X_i)$ that minimizes the cross-validated $R$-loss \eqref{eq:rloss}.
\end{algorithmic}
\label{alg:1}
\end{algorithm}

\subsection{Simulations under Varying Shared Support Sizes}
In a first step, we hypothesized that using the active support from the lasso model for $m(X_i)$ as pretraining information would improve the estimation of $\tau(X_i)$. The simulation here aims to test the ability of the pretrained $R$-lasso to improve on CATE estimation as we increase the size of the joint support. For ease of exposition, we demonstrate our approach in an easy toy example of a linear setting and generate data as follows. For more complex settings, we refer to \autoref{section:toc_power} as well as \autoref{section:non_linear}. All simulations are conducted in the statistical software \texttt{R} \citep{Rcore} and implemented using the packages \texttt{glmnet} \citep{friedman2010regularization} and \texttt{rlearner} \citep{nie2021quasi}.

Our experiments are conducted assuming a randomized trial setting. In this first experiment, we suppose $n$ units, for each we are given a binary treatment assignment $W_i \sim Binom(e(X_i))$, with $e(X_i) = 0.5$, $p$ covariates $X_i \sim \mathcal{N}(0, 1)^p$ and an outcome $Y_i = \mu_i + \varepsilon_i$. We define $\mu_i = \beta_0 + \sum_{j=1}^n X_{ij} \beta_j + W_i \cdot \tau(X_i)$, with $\tau(X_i) = \theta_0 + \sum_{j=1}^n X_{ij} \theta_j$, where $\beta_j \sim U(0.5, 1)$, $\theta_j \sim U(1, 2)$, and $\varepsilon_i \sim \mathcal{N}\left(0,\ \frac{Var(\mu)}{SNR} \right)$. We then introduce some sparsity and limit the size of the active support for $m(X_i)$ and $\tau(X_i)$ to $p/3$, respectively. We start with a fully disjoint support, which we gradually increase until the support is fully shared, i.e., equal to $p/3$. We evaluate performance in terms of expected mean-squared error (MSE) $\frac{1}{n} \sum_{i=1}^n[\tau(X_i) - \hat{\tau}(X_i)]^2$, which we normalize by the median MSE of the baseline $R$-lasso.

\begin{figure}[H]
  \centering
  \makebox[\textwidth][c]{\includegraphics[width=1\textwidth]{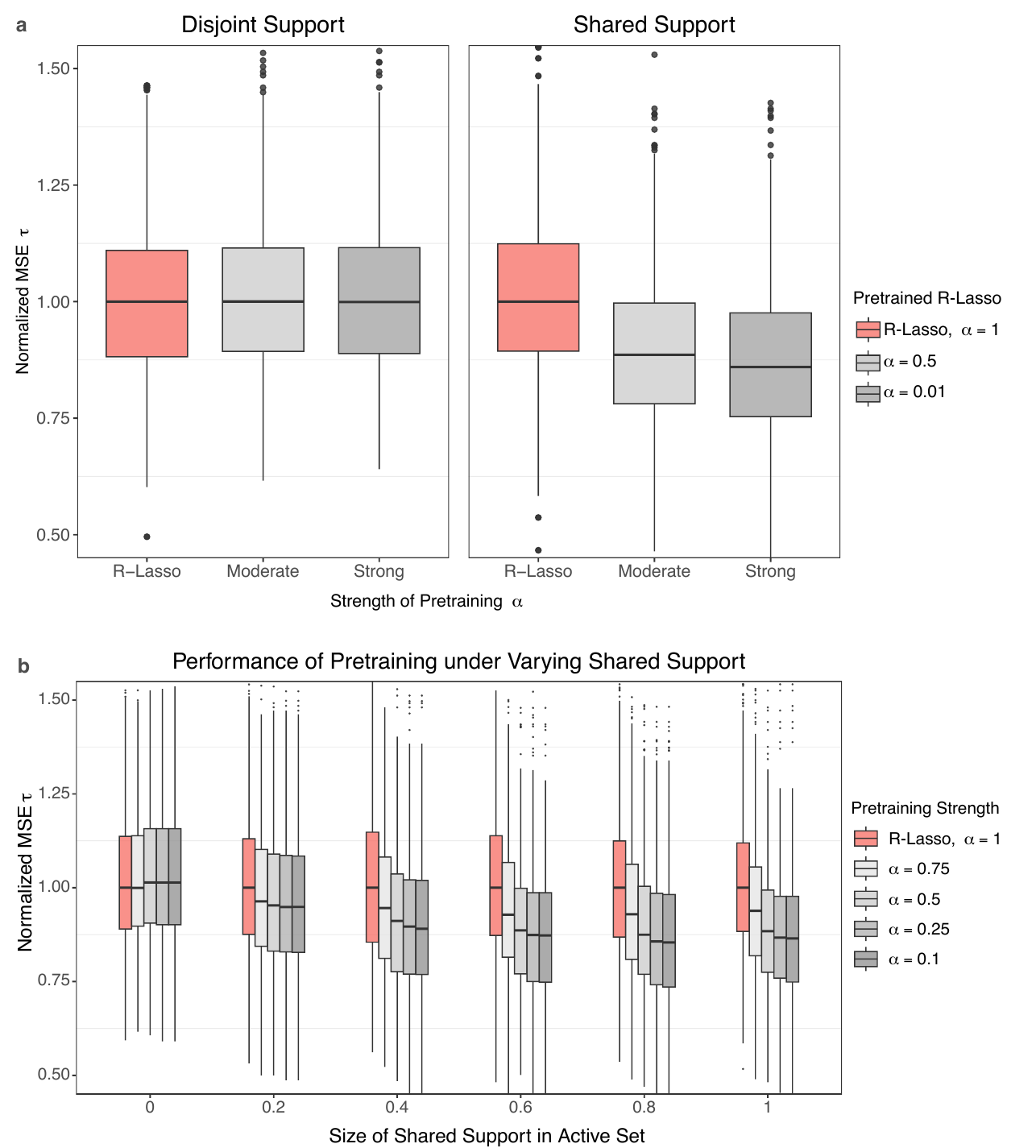}}
  \caption{\textbf{Effect of Pretraining on CATE Estimation in terms of MSE.} The figure shows the  performance of the $R$-lasso with (grey shades) and without pretraining (red) in a linear setting with $n = 500$, $p = 150$, and $SNR = 2$ and an active set of size $|S| = \frac{p}{3}$. Panel \textbf{a} shows the performance in two settings: with disjoint support between $m(X_i)$ and $\tau(X_i)$ (left) and shared support (right). The hyperparameter $\alpha$ denotes different pretraining strengths, where $\alpha = 1$ is equivalent to the $R$-lasso without pretraining and increasing pretraining strengths as $\alpha$ approaches 0.}
  \label{fig:fig_exp1_linear_rlasso}
\end{figure}

In our first experiment, we looked at the effect of pretraining the $R$-lasso in the two extreme cases, where the support is either fully disjoint or fully shared. Figure~\ref{fig:fig_exp1_linear_rlasso} shows the performance of the pretrained $R$-lasso in comparison to the baseline $R$-lasso in both settings and with different pretraining strengths (upper panel). Panel \textbf{b} shows the evolution of performance as the shared support increases from fully disjoint to fully joint support. In the absence of any overlap, pretraining shows no effect. With increasing shared support, the pretrained $R$-lasso achieves lower MSE, especially under strong pretraining (low $\alpha$).

\begin{figure}[H]
  \centering
  \makebox[\textwidth][c]{\includegraphics[width=1.0\textwidth]{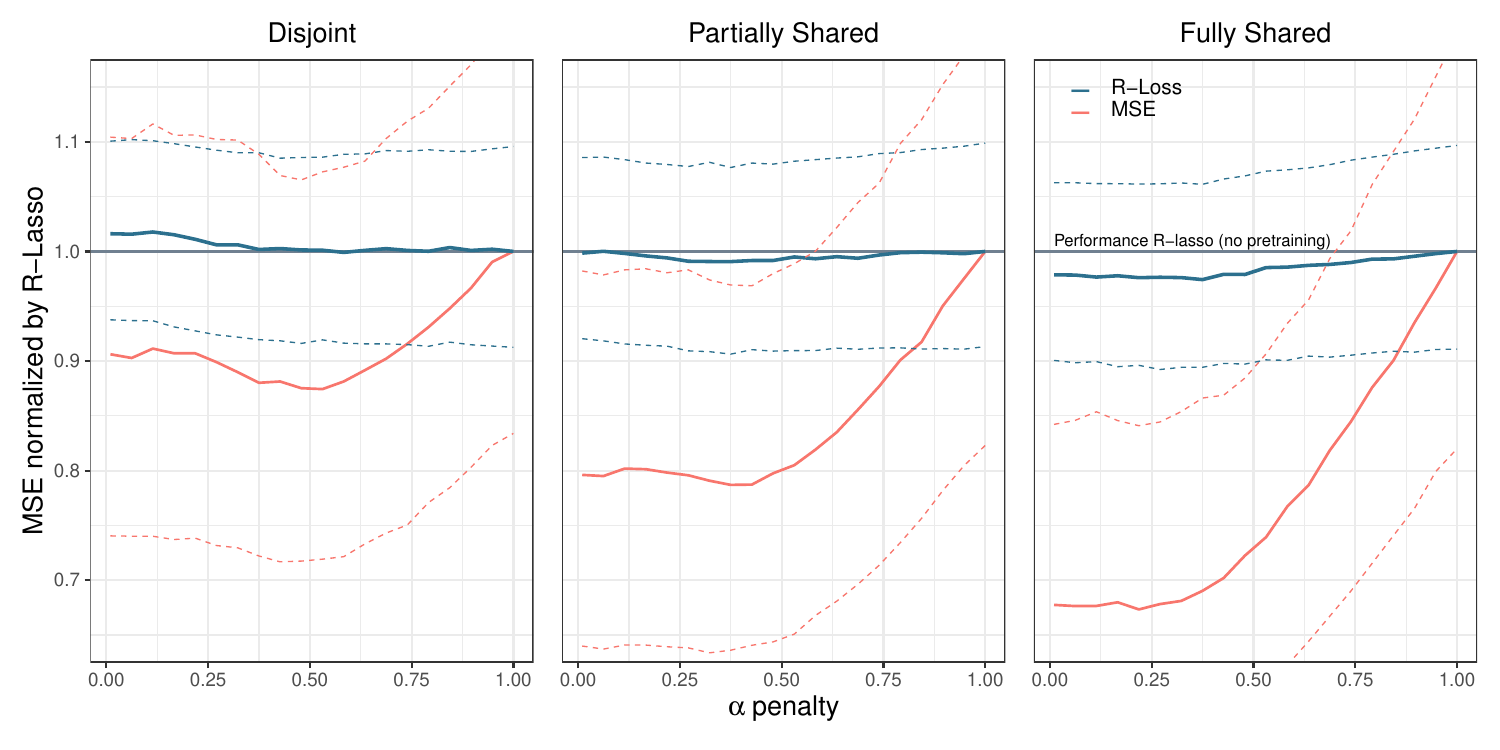}}
  \caption{ \textbf{Concordance of Training Set R-Loss and Test Set MSE.} The figure shows the  median performance (standardized by the median $R$-lasso without pretraining) oat different pretraining strengths of the $R$-lasso in a linear setting with $n = 500$, $p = 150$, $SNR = 2$, an active set of size $|S| = \frac{p}{3}$ and for $5000$ repetitions. The results are stratified by the size of shared support between prognostic and predictive factors: disjoint (left), partially shared (middle), and fully shared support (right panel). The minimum test set MSE and training set R-loss are concordant. In settings with stronger overlap, stronger pretraining (lower $\alpha$) is favorable. The solid line at 1.0 shows the $R$-lasso without pretraining ($\alpha = 1$) for comparison. Error bars denote the interquartile range between the 25th and 75th percentiles.}
  \label{fig:fig_rloss_mse_concordance}
\end{figure}

While the previous experiments showed a clear advantage for pretraining in settings with overlap, it requires the correct selection of the pretraining strength $w_j = \frac{1}{\alpha}$. Since the true $\tau(X_i)$ is not available to us, we need to rely on a surrogate loss and turn to the $R$-loss. Figure \ref{fig:fig_rloss_mse_concordance} shows the concordance between the $R$-loss estimated from the training set and the mean squared error (MSE) estimates on the test set. Encouragingly, the $R$-loss proves to be concordant with the MSE and allows us to pick $\alpha$ at which the test error is minimized. As we vary the size of overlap between prognostic and predictive factors, the pretrained $R$-lasso favors a lower $\alpha$ (stronger pretraining) as expected.

\subsection{Assessing Power to Detect Heterogeneity}\label{section:toc_power}
The end goal of fitting CATE models is to discover potential subgroups or strata as defined by our predictor variables $X_i$ that have average treatment effects that differ meaningfully from the overall population average. Quantifying their ability to achieve this goal provides valuable information. Recently proposed metrics useful for this task take inspiration from predictive analysis tools such as the ROC curve and include \citet{chernozhukov2018generic, radcliffe2007using, sverdrup2024qini, imai2023experimental, yadlowsky2024evaluating}. An intuitive approach to operationalize this is to take inspiration from classical machine learning, where we define a training set for fitting a CATE model, and evaluate its performance on a held-out test set. We can then rank individuals from highest to lowest predicted CATE. In the presence of heterogeneity, this should yield a decreasing quantile average treatment effect as we move from higher to lower quantiles. One way to operationalize this CATE-based ranking is to use the targeting operator characteristic (TOC) curve \citep{yadlowsky2024evaluating}, defined as 
\begin{equation}
\text{TOC}(q) = \mathbb{E} \left[Y_i(1) - Y_i(0) \mid \hat \tau(X_i) \geq F^{-1}_{\hat \tau(X_i)}(1 - q) \right]
                      - \mathbb{E} \left[ Y_i(1) - Y_i(0) \right].
\end{equation}
The area under the TOC (AUTOC) is a numeric measure of heterogeneity and allows us to evaluate whether and how well our CATE model detects heterogeneity against a sharp null hypothesis of no heterogeneity detected. One limitation of this approach is that train-test splits can be costly and suffer from random split sensitivity, especially in settings with small sample sizes. To overcome this limitation, we adopt a sequential cross-fold approach \citep{wager2024sequential}. 

\begin{figure}[H]
  \centering
  \makebox[\textwidth][c]{\includegraphics[width=1.1\textwidth]{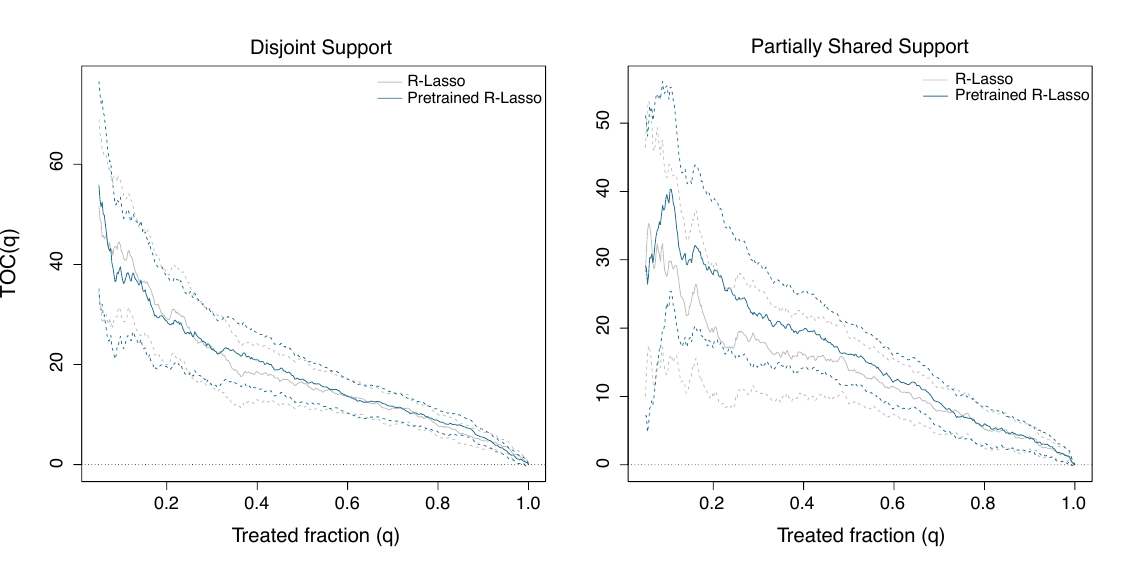}}
    \caption{\textbf{TOC-based Comparison of $R$-lasso with and without Pretraining.} The figure shows the Targeting Operator Characteristic curves for the two $R$-lasso configurations and in settings with (blue) and without (gray) shared support. Dashed lines denote 95\% confidence intervals based on the normal distribution. Earlier quintiles show higher average treatment effects.
    This experiment is conducted for a sparsity of $\kappa = 0.1$, $p = 500$, $n = 1000$, $SNR = 2$ and a shared support size of $p \cdot \kappa \cdot 0.7$.}
    \label{fig:fig_toc_pretrained}
\end{figure}

\begin{figure}[H]
  \centering
  \makebox[\textwidth][c]{\includegraphics[width=1.1\textwidth]{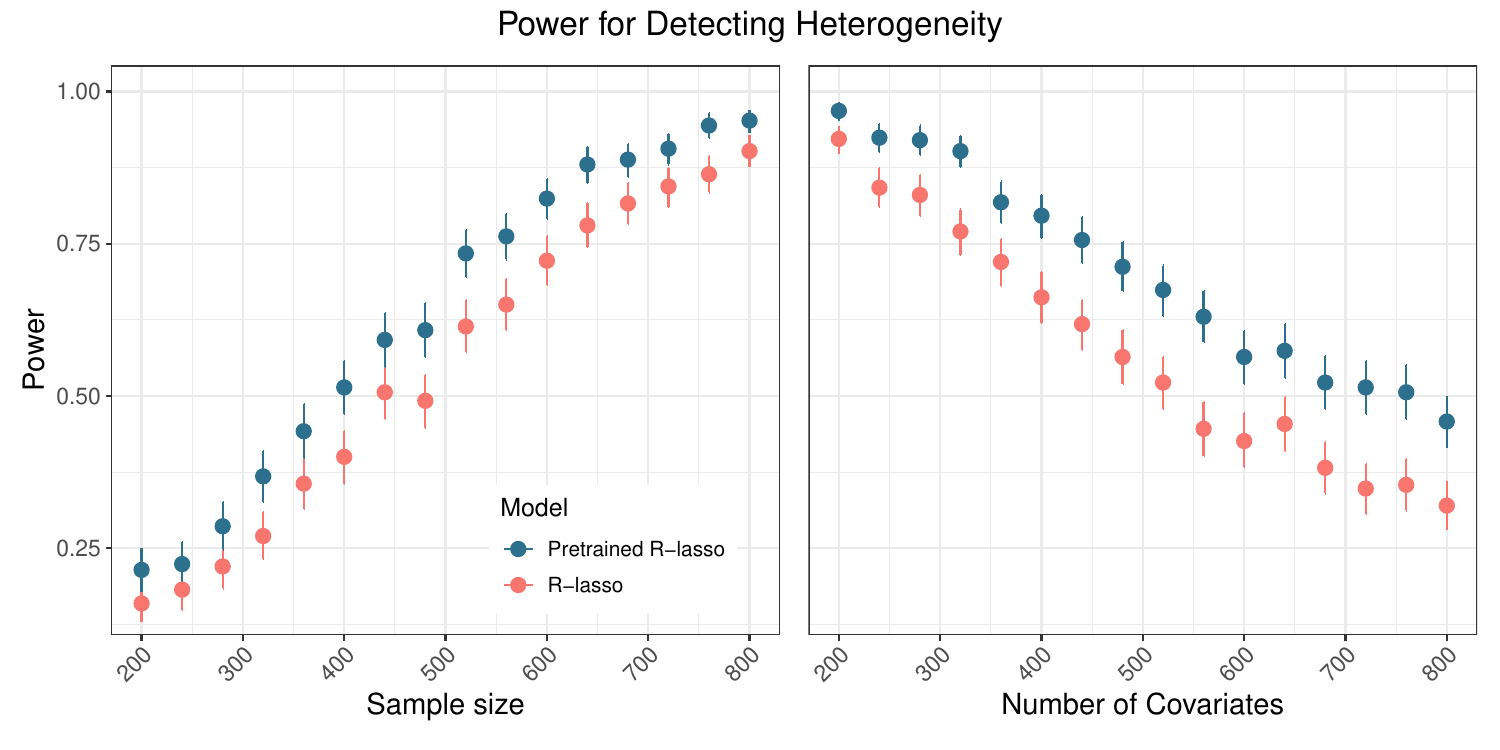}}
    \caption{\textbf{Power in Detecting Heterogeneous Treatment Effects.} The figure shows the power of detecting heterogeneity for the pretrained and default $R$-lasso. The experiment is repeated 200 times under 5-fold sequential cross-validation and sparsity of $\kappa = 0.1$, joint support size of $p \cdot \kappa \cdot 0.7$, $SNR = 0.5$, while varying $n$ (left panel, $p = 200)$ or varying $p$ (right panel, $n = 800$). Error bars denote 95\% bootstrapped confidence interval.}
      \label{fig:fig_power_HTE}
\end{figure}

 Using the TOC curve, we can assess the power at which our CATE model detects heterogeneity. We hypothesized that in the presence of shared support, the pretrained $R$-lasso should be more powerful in detecting heterogeneity than without pretraining. For this experiment, we slightly tweak our DGP to a more challenging setting and reduce the support size to $p/10$, half of which is shared between $m(X_i)$ and $\tau(X_i)$, and the rest is disjoint. We further introduce correlation among our covariates $X \sim \mathcal{N}_p(\mathbf{0}, \Sigma), \text{where } \Sigma \text{ is a Toeplitz matrix with entries } \Sigma_{ij} = 0.5^{|i - j|}$ and reduce the SNR to 0.5. Figure \ref{fig:fig_toc_pretrained} shows the TOC curve in two configurations: fully disjoint support and partially shared support. In the presence of overlap, the pretrained $R$-lasso shows a superior AUTOC as well as high power in detecting heterogeneity as we vary the sample size $n$ and the number of covariates $p$ (Figure~\ref{fig:fig_power_HTE}).

\subsection{Sparsity and Reliable Support Recovery: uniR-Lasso}

As demonstrated in the previous section, the effectiveness of our pretraining strategy depends both on the presence of overlap and the $m(X_i)$ model’s ability to accurately detect it. The lasso has been shown to be less reliable for this task in the presence of highly correlated features and false positives in the active set \citep{su2017false}. To address this concern, \citet{chatterjee2025univariate} recently developed a univariate-guided sparse regression termed ``UniLasso'' that achieves higher sparsity by constraining the coefficients to have the same sign and reflecting their absolute magnitudes in univariate regression. 

Sparsity is also a desirable property in our setting as it will result in more covariates being penalized. Specifically, it will result in stronger, more precise pretraining when fitting the model for $\tau(X_i)$. In addition, this model also achieves higher interpretability in high-dimensional settings with low signal, which might be desirable in many settings. As highlighted by \citet{candes2018panning}, identifying a subset of relevant covariates, while controlling false discovery rates (FDR), is highly valuable in many areas \citep{storey2003statistical}. For example, in biology, the reliable identification of a single driver of heterogeneity can help discover a larger regulatory network that modifies treatment effects.

Figure \ref{fig:unilasso} shows the $R$-lasso and Uni$R$-Lasso in a side-by-side comparison in terms of MSE, size of the active set, FDR, and the Jaccard index for the DGP described in \autoref{section:toc_power}. The pretraining strength for both models is determined using the $R$-loss on the test data. The results show comparable performance in terms of MSE for both models. However, uni$R$-lasso shows superior control of FDR and a better support recovery as measured by the Jaccard Index. Both models further benefit from pretraining proportionally to the size of shared support.  

\begin{figure}[H]
  \centering
  \makebox[\textwidth][c]{\includegraphics[width=1.05\textwidth]{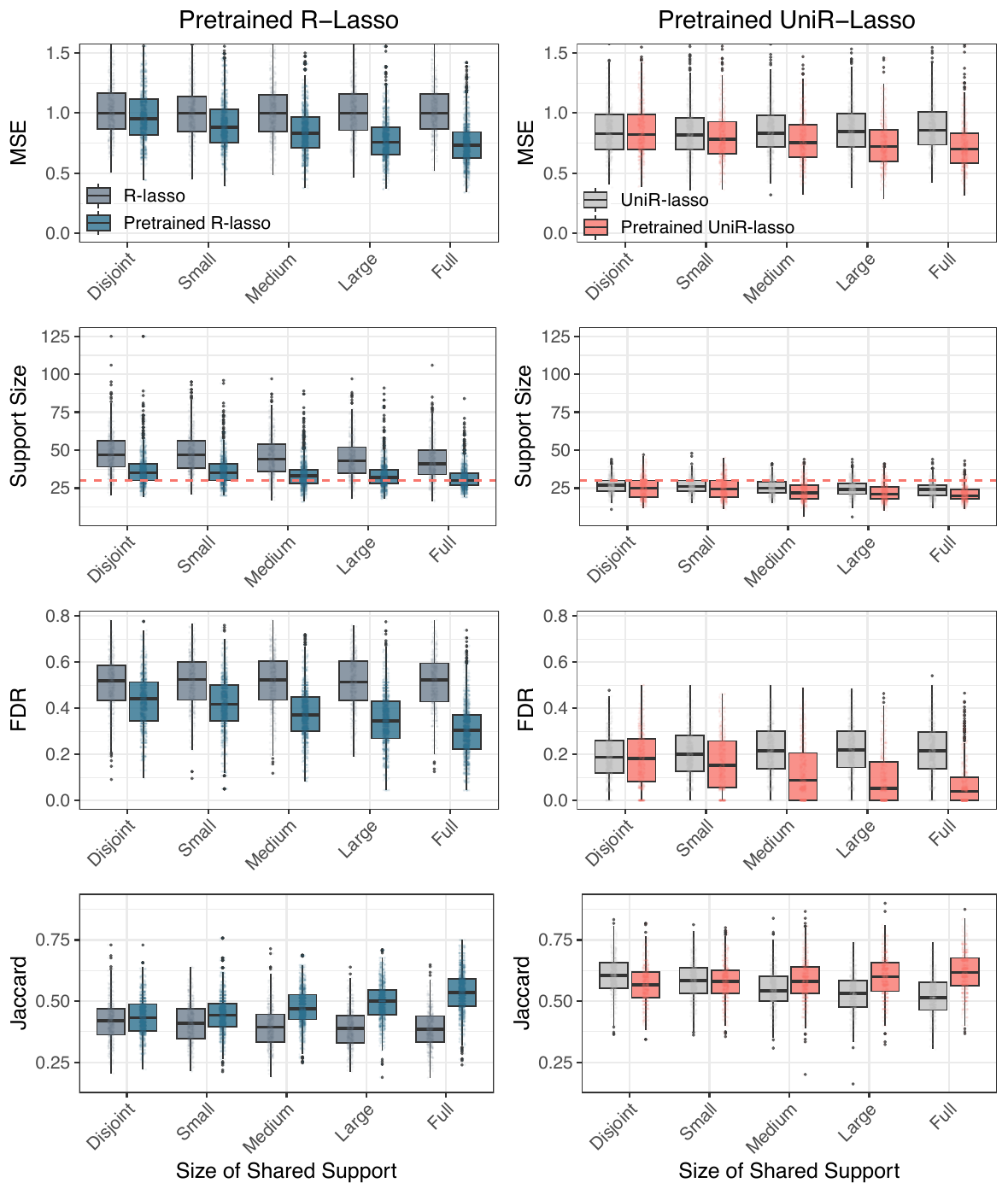}}
    \caption{\textbf{Comparison of Pretraining for $R$-lasso and Uni$R$-lasso} The figure shows the performance in terms of mean squared error (MSE), Support Size (red dashed indicates the true support), False Discovery Rate (FDR), and Jaccard Index. The MSE is normalized by the $R$-lasso for each shared support size setting. The experiment is conducted using the DGP described \autoref{section:toc_power}, with $p = 300$, $n = 500$ and $SNR = 2$. The strength of pretraining $\alpha$ is determined using the R-loss for both learners.}
  \label{fig:unilasso}
\end{figure}

\section{Leveraging Side Information in Non-Linear Settings} \label{section:non_linear}
In many applications, the mean outcome and CATE functions might involve complex interactions and non-linearities that the linear lasso outlined in \autoref{section:rlassolinear} may not capture. A straightforward solution to continue using the lasso without modifying Algorithm \ref{alg:1} is to substitute the feature matrix $X_i$ with $\psi(X_i)$ where $\psi: \mathcal{X} \mapsto \mathbb{R}^d$ is some some featurization that maps the covariates to a richer $d$-dimensional space. $\psi$ could, for example, be a collection of polynomial basis functions. With a rich set of basis functions, this approach can sufficiently capture complicated CATE or mean outcome functions (e.g., \citet{chen2007large}). While a wide variety of basis expansions exist and the optimal choice is often application-specific \citep{friedman1991multivariate}, here we concentrate on gradient boosted trees to generate these basis expansions \citep{chen2016xgboost, friedman2001greedy} and use the statistical software package \texttt{xgboost} \citep{chen2016xgboost}. Gradient boosting is an additive model that fits a sequence of weighted trees or weak learners. We explore its use in the $R$-learner for $M = 100$ and $500$ boosting rounds and a tree depth of 1 (stump). This results in a total of $M$ trees (and thus new features), each of which performs a single split. As we increase the tree depth and number of boosting rounds, the transformed feature space is able to capture more complex relations, but also suffers from higher dimensions. To perform hyperparameter tuning, i.e., select the strength of pretraining and the number of boosting rounds, we use the $R$-loss. By concatenating this new feature matrix with our original one, the lasso can now pick from the original and transformed features. This is particularly useful in partially linear settings, where estimating $m(X_i)$ and $\tau(X_i)$ might require a model capable of handling both linear and non-linear components simultaneously.  Depending on whether we expect $m(X_i)$ and $\tau(X_i)$ to differ substantially in their functional forms, we can generate distinct sets of basis functions tailored to each prediction task in the $R$-learner.

\begin{figure}[H]
  \centering
  \makebox[\textwidth][c]{\includegraphics[width=1.0\textwidth]{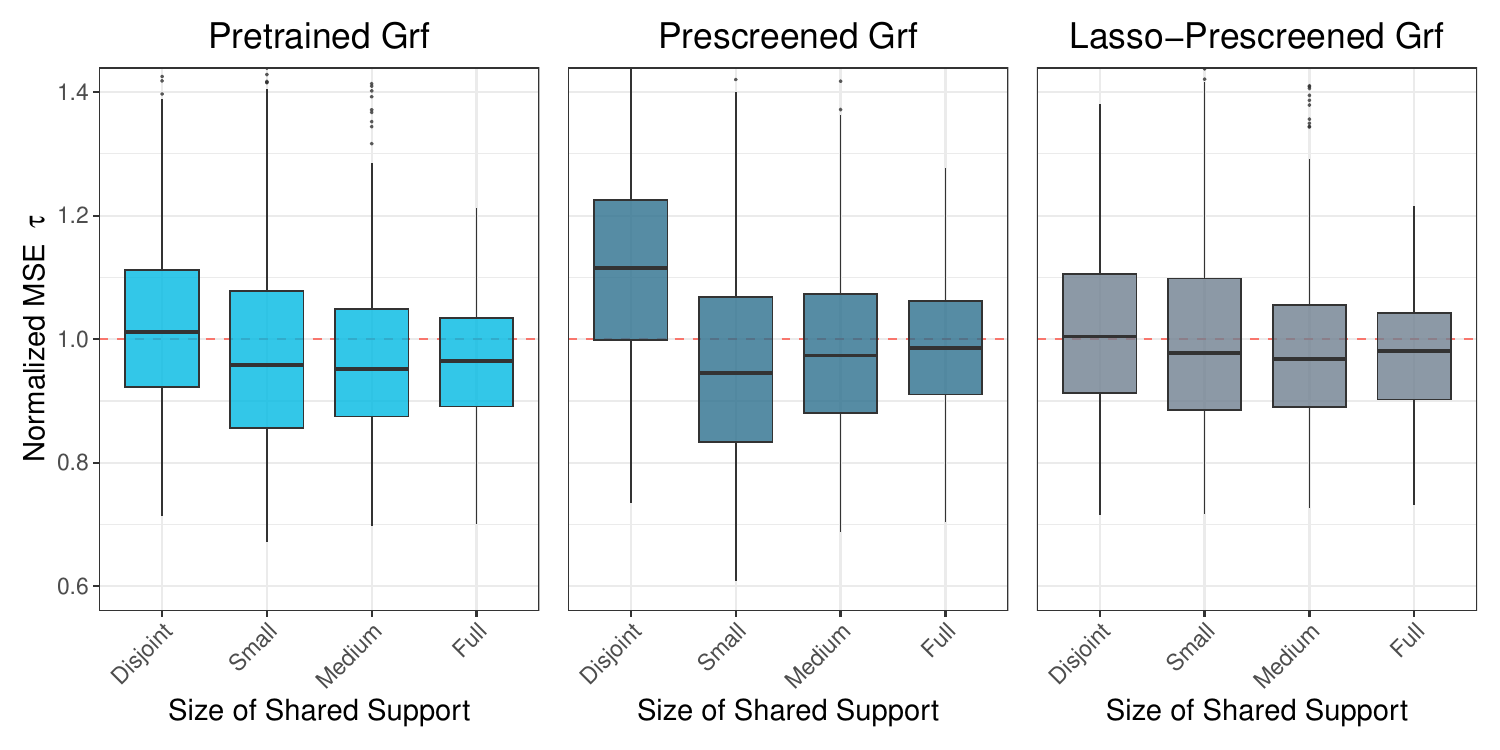}}
    \caption{\textbf{Effect of Pretraining Strategies for Generalized Random Forests (GRF).} The figure shows the performance in terms of mean squared error (MSE), normalized by the baseline GRF, in three different configurations: when using weighted feature sampling based on $m(x)$ (``Pretrained  GRF'', left panel), prior variable prescreening and removal (``Prescreened GRF'', middle panel) based on importance scores, and prescreening based on the lasso (``Lasso-Prescreened GRF'', right panel). The experiment is conducted using the DGP described in \autoref{section:toc_power}, with $p = 150$, $n = 500$, and $SNR = 2$. The degree to which importance scores from $m(x)$ are translated into differential feature sampling/screening is determined by the $R$-loss.}
  \label{fig:fig_grf_pretraining}
\end{figure}

The lasso solution to sharing side information leverages the lasso's adaptive regularization feature, but similar logic can be applied to other learners, such as random forests. Causal forests \citep{athey2019generalized} are forest-based instantiations of the $R$-learner that uses regression forests for $m(X_i)$ and $e(X_i)$, and then fits $\tau(X_i)$ using a forest-weighted version of \eqref{eq:empirical_rmin} designed to express heterogeneity in the CATE.  As trees rely on split criteria rather than shrinking coefficients as in the lasso, we require a new strategy that allows us to leverage side information. An effective way to improve random forests is to use adaptive feature-weighting (e.g., \citet{basu2018iterative}). Adapted to our setting, this procedure consists of fitting a random forest for $m(X_i)$, deriving the feature importance score for each covariate, and, finally, sampling each feature in $\tau(X_i)$ with a probability that reflects its importance in $m(X_i)$. As for the lasso, we can introduce a hyperparameter that determines the strength at which the feature importance gets translated into differential feature sampling. We refer to this approach as ``Pretrained GRF''. A related approach is to use $m(X_i)$ for selecting/discarding features based on their importance. If a given feature's importance is below a certain threshold in grf (``prescreened GRF''), it is dropped or sampled with higher probability based on its coefficients from a lasso model (``lasso-prescreened GRF''), when estimating the $\tau(X_i)$ model. Figure \ref{fig:fig_grf_pretraining} shows the performance of these modified CATE estimators using the DGP described in \autoref{section:toc_power} using the \texttt{grf} package \citep{GRF}.

Finally, we test these models against the pretrained $R$-lasso in combination with basis expansions generated by a gradient boosted tree. To test these approaches, we modified four DGPs developed by \citet{nie2021quasi}. We consider four setups A-D, each varying in the complexity of the treatment effect $\tau(X_i)$ and the main effects function $\mu_{(0)}(X_i)$. Setups A-C assume a randomized controlled trial with $e(X_i) = 0.5$, while the propensities in D are given by $e(X_i) = 1/(1 + \exp(-X_{i1}) + \exp(-X_{i2}))$. Covariates are generated as $X \sim \mathcal{N}(0, 1)^{n \times p}$. Setup A  features a non-linear baseline main effect given by $\mu_{(0)}(X_i) = \sin(\cdot X_{i1} X_{i2}) + 2(X_{i3} - 0.5)^2 + X_{i4}^3 + 0.5 X_{i5}^3$ and a linear treatment effect, defined as $\tau(X_i) = (\sum_{j=1}^{5} X_{ij}) / 5$. Setup B introduces high-dimensional interactions: $\mu_{(0)}(X_i) = \max\{0, X_{i1}^2 + X_{i2}^2, X_{i3}^2\} + \max\{0, X_{i4}^2 + X_{i5}^2\} + X_{i6}^2 + X_{i7}^2$ and a more complex, non-linear treatment effect: $\tau(X_i) = X_{i1}^2 + \log(1 + \sum_{j=2}^{5} e^{X_{ij}})$. Setup C flips the structure of setup A: the baseline function is linear, and the treatment effect is complex. The outcome model is $\mu_{(0)}(X_i) = (\sum_{j=1}^{5} X_{ij}) / 5 + 0.8(X_{i6} + X_{i7})$ and the treatment effect is defined as $\tau(X_i) = \max\{0, \sum_{j=1}^{3} X_{ij} + \max\{0, X_{i4}^2 + X_{i5}^2\} + X_{i6}^2 + X_{i7}^2 + 0.5(X_{i8} + X_{i9})$. Finally, Setup D presents a baseline outcome that combines piecewise linearity and quadratic effects: $\mu_{(0)}(X_i) = \left[\max\{0, \sum_{j=1}^{3} X_{ij}\} + \max\{0, X_{i4} + X_{i5}\}\right]/2 + X_{i6}^2 + X_{i7}^2$. The treatment effect is composed of contrasting additive and subtractive components: $\tau(X_i) = \max\{0, \sum_{j=1}^{3} X_{ij}\} - \max\{0, X_{i4} + X_{i5}\} + X_{i9}^2 + X_{i10}^2$. 

\begin{figure}[H]
  \centering
  \makebox[\textwidth][c]{\includegraphics[width=1.0\textwidth]{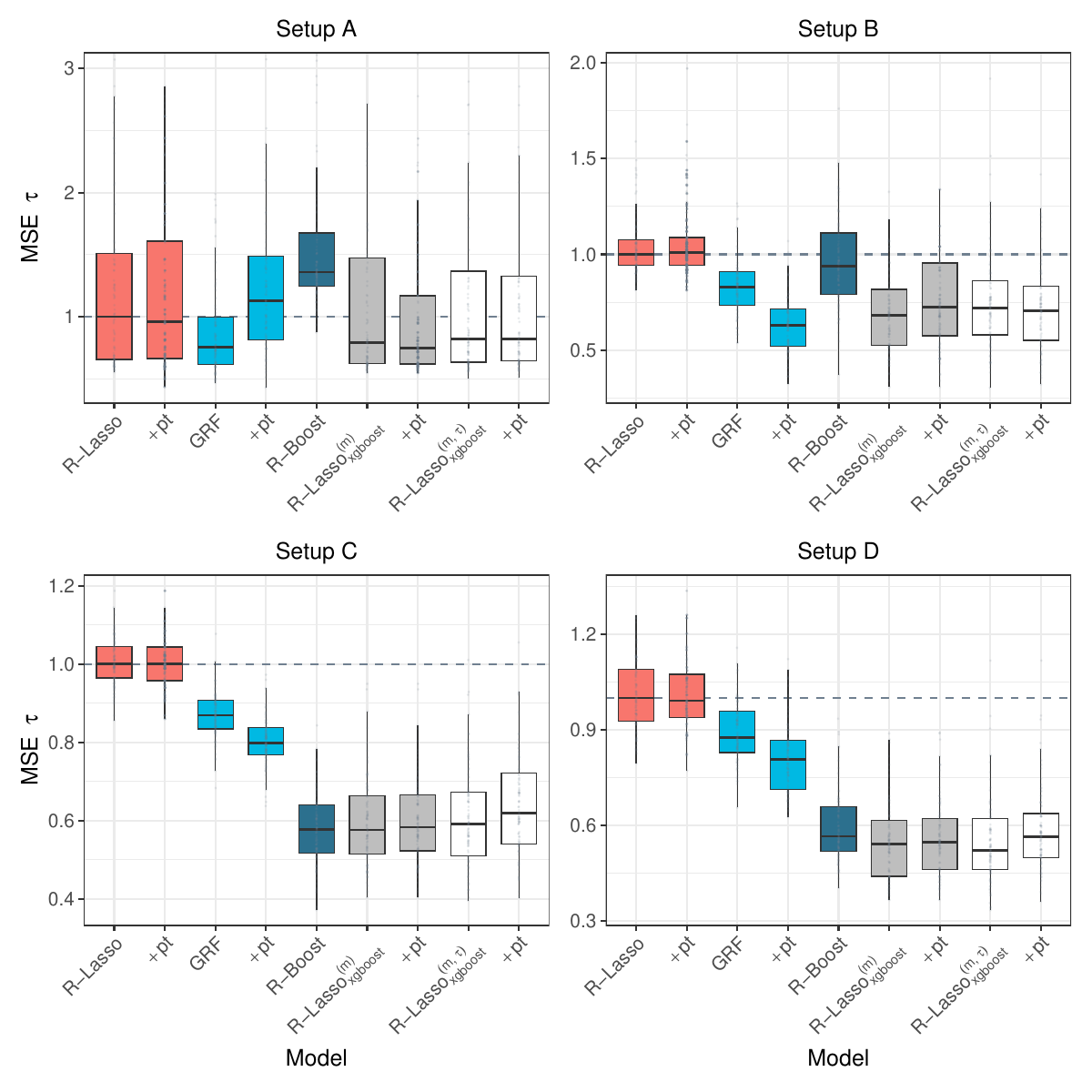}}
    \caption{\textbf{Comparison of CATE Estimators in Non-linear Settings.} The figure shows the performance in terms of mean squared error (MSE), normalized by the baseline $R$-lasso. The experiment is conducted using the DGPs described for setups A-D, with $p = 30$, $n = 1000$, and $SNR = 2$. The degree of pretraining in each model is determined by the $R$-loss. The $R$-lasso models with subscripts $xgboost$ denote models that take basis functions generated from the fit of an xgboost on $m(X_i)$ as input. The model with superscript $(m, \tau)$ leverages basis functions generated from both fits on $m(X_i)$ and $\tau(X_i)$. The dashed line indicates the performance of the $R$-lasso by which all MSEs are standardized.}
\label{fig:nonlinear_setups}
\end{figure}

Figure \ref{fig:nonlinear_setups} shows the performance of all models, including the (pretrained) $R$-lasso, (pretrained) GRF, and $R$-Boost. We further include two $R$-lasso models (with and without pretraining) that take basis expansions generated from xgboost, concatenated with the original design matrix, as input: the first model configuration uses the basis functions generated from the fit on $m(X_i)$; the second leverages basis functions generated from fits on both $m(X_i)$ and $\tau(X_i)$, but shields the basis functions generated from the latter from any pretraining penalties. We report results for all combinations of $n \in \{500, 1000\}$, $p \in \{15, 30\}$ and SNR $\in \{0.5, 2\}$. 

\section{Discussion}
In this paper, we present a new approach to pretrain CATE models that minimize a causal loss function. Our pretraining strategy is motivated by a real-world phenomenon that gives rise to distinct structural patterns in the data. In many settings, factors associated with prognosis are also predictive of high treatment effects. The intuition behind our approach is that rather than treating the estimates from the mean outcome model and the propensity model as mere nuisance parameters, we leverage their supports as side information. This allows us to improve on our CATE estimation and synergize independent prediction tasks in the $R$-learner. While our exposition focused mostly on the lasso, we also demonstrated how our pretraining approach can be extended to settings with multicollinearity using the uni$R$-lasso or to non-linear settings via basis functions and generalized random forests. Employing this pretraining strategy in the $R$-learner resulted in lower error rates, higher power in detecting heterogeneity, and lower false discovery rates, which is particularly relevant in fields like biomarker discovery.

One limitation of our pretraining approach is that it offers no performance benefit in scenarios with minimal or no overlap between predictive and prognostic factors. In such settings, it is possible that pretraining might lead to over-regularization of the wrong covariates and prioritize strong predictors in the shared support over disjoint weak predictors in the treatment effect function. This property might not be favorable, especially when treatment effects are spread out across a few strong and many weak predictors. Another limitation is the dependence on the $R$-loss for correct choice of $\alpha$ and other hyperparameters: if the estimation error for nuisance parameters is large, the $R$-loss becomes less reliable for such hyperparameters.   

A number of extensions of our approach are left open. The proposed forms of supervised pretraining used in this work are demonstrated for settings where researchers have access to a single dataset. However, in many settings, it might be valuable to pretrain CATE models on an additional external dataset. Our approach can be extended to those settings as it is possible to pretrain the models for $m(X_i)$, $e(X_i)$, and $\tau(X_i)$ using an external dataset. Similarly, researchers may have access to a large dataset spanning multiple cancer types that share some common features. In this case, the $R$-Learner can first be trained on the full dataset, with penalties from each prediction task transferred to refine a second, cancer-specific $R$-Learner (e.g., for breast cancer or subtypes defined by biomarkers such as HER2). Pretraining can help obtain targeted insights, a task often made difficult by low signal, small sample sizes, and high dimensionality. Finally, our approach to incorporating side information relied on a covariate support set. Exploring alternative strategies that directly leverage coefficient estimates or other model components, such as baseline risk probabilities (perhaps via Bayesian approaches \citep{hill2011bayesian}), presents interesting directions for future work.

\section*{Acknowledgments}
We are grateful to Erin Craig, Trevor Hastie, Guido Imbens, Ying Lu, Jonathan Taylor, Lu Tian, and Stefan Wager, as well as seminar participants at Stanford University, for valuable comments and feedback.

R.T. was supported by the NIH (5R01EB001988-16) and the NSF (19DMS1208164).

\section*{Declaration of interest statement}
The authors declare no conflict of interest.

% -----------------------------------------------------------------------------------
%	BIBLIOGRAPHY
%----------------------------------------------------------------------------------------

\bibliography{refs} 
\bibliographystyle{agsm} 
\newpage
\clearpage

\appendix
\section{Supplement}

\begin{figure}[H]
  \centering
  \makebox[\textwidth][c]{\includegraphics[width=1.0\textwidth]{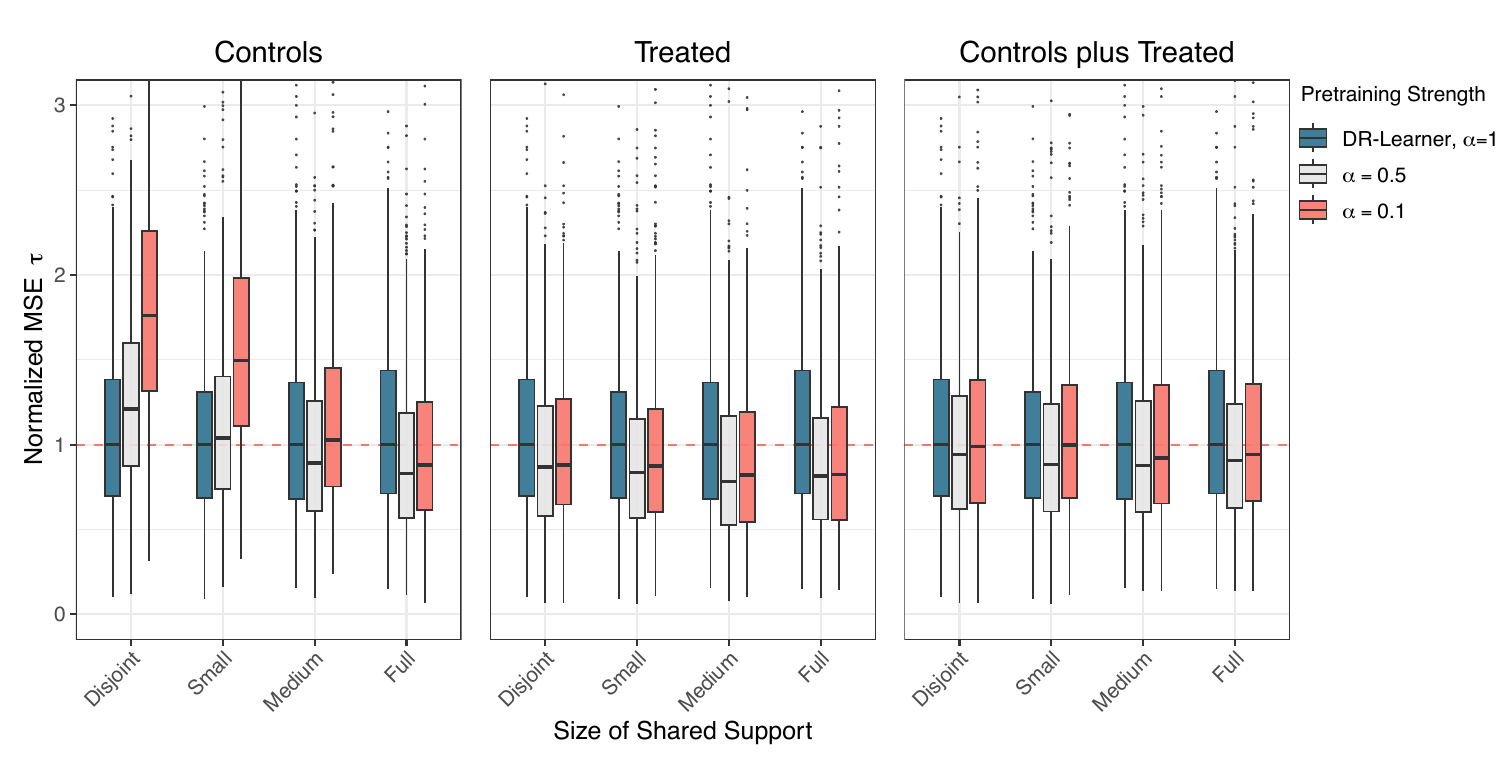}}
    \caption{\textbf{Pretraining of the DR-Learner.} The barplots shows model the performance in terms of mean squared error (MSE) when the CATE function in DR-Learner is pretrained using the adaptive lasso and side information from the controls through $\mu_0(x)$, the treated units through $\mu_1(x)$ or both. The MSE is normalized by the $DR$-lasso with no pretraining ($\alpha$ = 1) for each support size. The experiment is conducted using the DGP described \autoref{section:toc_power}, with $p = 60$, $n = 500$ and $SNR = 2$.}
  \label{fig:unilasso}
\end{figure}

\newgeometry{margin=1.0cm} 
\begin{landscape}

\begin{table}[h]
\thispagestyle{empty}
\centering
\caption{\textbf{Comparison of R-lasso and UniR-lasso (with Pretraining) in Linear Setting with Sparsity k = 0.1}}
\centering
\fontsize{10}{12}\selectfont
\begin{tabular}[t]{rrrllll}
\toprule
\multicolumn{3}{c}{ } & \multicolumn{2}{c}{R-lasso} & \multicolumn{2}{c}{UniR-lasso} \\
\cmidrule(l{3pt}r{3pt}){4-5} \cmidrule(l{3pt}r{3pt}){6-7}
\multicolumn{1}{c}{n} & \multicolumn{1}{c}{p} & \multicolumn{1}{c}{SNR} & \multicolumn{1}{c}{Base} & \multicolumn{1}{c}{+pt} & \multicolumn{1}{c}{Base} & \multicolumn{1}{c}{+pt}\\ 
\midrule
\multicolumn{7}{l}{\textit{Mean Squared Error}}\\
\midrule
300 & 50 & 0.5 & 17.48 (17.12, 17.82) & 14.33 (14.16, 14.48) & 13.48 (13.18, 13.78) & \textbf{12.31 (12.04, 12.59)}\\
300 & 50 & 2.0 & 5.32 (5.22, 5.42) & 4.08 (4.03, 4.13) & 3.29 (3.21, 3.37) & \textbf{2.97 (2.89, 3.03)}\\
300 & 300 & 0.5 & 317.54 (314.58, 320.70) & 295.88 (294.26, 297.67) & 309.32 (306.55, 312.09) & \textbf{291.98 (289.58, 294.20)}\\
300 & 300 & 2.0 & 166.93 (165.21, 168.88) & \textbf{142.01 (141.14, 142.83)} & 156.55 (154.97, 158.17) & 144.85 (143.56, 146.22)\\
500 & 50 & 0.5 & 10.76 (10.57, 10.97) & 8.50 (8.41, 8.59) & 7.79 (7.63, 7.98) & \textbf{6.93 (6.78, 7.08)}\\

500 & 50 & 2.0 & 3.03 (2.96, 3.09) & 2.27 (2.24, 2.30) & 1.70 (1.66, 1.75) & \textbf{1.57 (1.53, 1.61)}\\
500 & 300 & 0.5 & 239.49 (236.99, 241.86) & \textbf{211.28 (209.95, 212.61)} & 233.15 (230.81, 235.58) & 217.83 (215.66, 220.02)\\
500 & 300 & 2.0 & 103.96 (102.97, 104.96) & 86.05 (85.56, 86.57) & 88.20 (87.20, 89.18) & \textbf{79.63 (78.76, 80.48)}\\
\midrule \multicolumn{7}{l}{\textit{Jaccard Index}} \\ \midrule
300 & 50 & 0.5 & 0.41 (0.40, 0.42) & 0.52 (0.51, 0.52) & 0.56 (0.55, 0.56) & \textbf{0.58 (0.58, 0.59)}\\
300 & 50 & 2.0 & 0.50 (0.49, 0.50) & 0.65 (0.64, 0.65) & 0.83 (0.82, 0.83) & \textbf{0.87 (0.87, 0.88)}\\

300 & 300 & 0.5 & 0.13 (0.13, 0.13) & \textbf{0.16 (0.16, 0.16)} & 0.13 (0.13, 0.13) & 0.13 (0.13, 0.14)\\
300 & 300 & 2.0 & 0.32 (0.32, 0.32) & 0.38 (0.38, 0.38) & 0.38 (0.38, 0.38) & \textbf{0.40 (0.40, 0.40)}\\
500 & 50 & 0.5 & 0.47 (0.46, 0.48) & 0.61 (0.60, 0.61) & 0.70 (0.69, 0.71) & \textbf{0.74 (0.73, 0.75)}\\
500 & 50 & 2.0 & 0.50 (0.49, 0.51) & 0.66 (0.66, 0.67) & 0.86 (0.85, 0.86) & \textbf{0.89 (0.89, 0.90)}\\
500 & 300 & 0.5 & 0.23 (0.22, 0.23) & \textbf{0.28 (0.27, 0.28)} & 0.24 (0.23, 0.24) & 0.26 (0.25, 0.26)\\

500 & 300 & 2.0 & 0.40 (0.40, 0.40) & 0.48 (0.48, 0.48) & 0.56 (0.55, 0.56) & \textbf{0.59 (0.59, 0.59)}\\
\midrule \multicolumn{7}{l}{\textit{False Discovery Rate}} \\ \midrule
300 & 50 & 0.5 & 0.47 (0.46, 0.48) & 0.33 (0.32, 0.33) & 0.19 (0.18, 0.19) & \textbf{0.14 (0.13, 0.14)}\\
300 & 50 & 2.0 & 0.49 (0.49, 0.50) & 0.34 (0.33, 0.34) & 0.15 (0.14, 0.15) & \textbf{0.10 (0.10, 0.11)}\\
300 & 300 & 0.5 & 0.44 (0.43, 0.46) & 0.36 (0.35, 0.36) & 0.39 (0.38, 0.40) & \textbf{0.18 (0.18, 0.19)}\\
300 & 300 & 2.0 & 0.52 (0.52, 0.53) & 0.40 (0.39, 0.40) & 0.27 (0.26, 0.27) & \textbf{0.14 (0.13, 0.14)}\\

500 & 50 & 0.5 & 0.48 (0.47, 0.49) & 0.32 (0.32, 0.33) & 0.16 (0.16, 0.17) & \textbf{0.11 (0.11, 0.12)}\\
500 & 50 & 2.0 & 0.50 (0.49, 0.51) & 0.33 (0.33, 0.34) & 0.14 (0.13, 0.15) & \textbf{0.10 (0.10, 0.11)}\\
500 & 300 & 0.5 & 0.49 (0.48, 0.50) & 0.37 (0.37, 0.38) & 0.32 (0.32, 0.33) & \textbf{0.18 (0.18, 0.19)}\\
500 & 300 & 2.0 & 0.51 (0.51, 0.52) & 0.37 (0.37, 0.38) & 0.21 (0.21, 0.21) & \textbf{0.13 (0.12, 0.13)}\\
\midrule \multicolumn{7}{l}{\textit{Support Size}} \\ \midrule
300 & 50 & 0.5 & 8.94 (8.73, 9.15) & 6.41 (6.32, 6.49) & 4.40 (4.33, 4.48) & \textbf{4.18 (4.11, 4.26)}\\

300 & 50 & 2.0 & 11.47 (11.28, 11.67) & 8.29 (8.21, 8.36) & 5.96 (5.89, 6.02) & \textbf{5.64 (5.58, 5.69)}\\
300 & 300 & 0.5 & 15.79 (15.18, 16.36) & 12.10 (11.87, 12.35) & 8.70 (8.46, 8.93) & \textbf{6.57 (6.38, 6.79)}\\
300 & 300 & 2.0 & 36.82 (36.22, 37.42) & 27.67 (27.43, 27.89) & 19.21 (18.98, 19.47) & \textbf{16.02 (15.83, 16.21)}\\
500 & 50 & 0.5 & 10.16 (9.98, 10.37) & 7.34 (7.27, 7.42) & 5.22 (5.15, 5.29) & \textbf{4.93 (4.87, 4.99)}\\
500 & 50 & 2.0 & 11.72 (11.52, 11.92) & 8.41 (8.33, 8.48) & 6.05 (5.99, 6.10) & \textbf{5.78 (5.72, 5.83)}\\

500 & 300 & 0.5 & 25.17 (24.59, 25.73) & 19.40 (19.15, 19.65) & 13.18 (12.92, 13.44) & \textbf{12.13 (11.84, 12.42)}\\
500 & 300 & 2.0 & 46.11 (45.58, 46.66) & 34.16 (33.94, 34.40) & 25.66 (25.45, 25.86) & \textbf{23.42 (23.16, 23.67)}\\
\bottomrule
\end{tabular}
\end{table}

\begin{table}[h]
\thispagestyle{empty}
\centering
\caption{\textbf{Comparison of CATE Estimators in Non-linear Settings: Test Set MSE}}
\centering
\begin{tabular}[t]{lrrrlllllllll}
\toprule
\multicolumn{4}{c}{ } & \multicolumn{6}{c}{R-lasso} & \multicolumn{1}{c}{R-boost} & \multicolumn{2}{c}{GRF} \\
\cmidrule(l{3pt}r{3pt}){5-10} \cmidrule(l{3pt}r{3pt}){11-11} \cmidrule(l{3pt}r{3pt}){12-13}
\multicolumn{1}{c}{Setup} & \multicolumn{1}{c}{n} & \multicolumn{1}{c}{p} & \multicolumn{1}{c}{SNR} & \multicolumn{1}{c}{Base} & \multicolumn{1}{c}{+pt} & \multicolumn{1}{c}{+xgb(m)} & \multicolumn{1}{c}{+pt} & \multicolumn{1}{c}{+xgb(m, tau)} & \multicolumn{1}{c}{+pt} & \multicolumn{1}{c}{} & \multicolumn{1}{c}{Base} & \multicolumn{1}{c}{+pt} \\
\midrule
A & 500 & 15 & 0.5 & 1.19 & 0.93 & 0.59 & \textbf{0.52} & 0.73 & 0.76 & 0.58 & 0.88 & 0.91\\
A & 500 & 15 & 2.0 & 0.51 & 0.43 & 0.35 & \textbf{0.29} & 0.43 & 0.45 & 0.49 & 0.40 & 0.43\\
A & 500 & 30 & 0.5 & 0.96 & 1.04 & 0.62 & \textbf{0.49} & 1.13 & 0.94 & 0.75 & 0.79 & 0.86\\
A & 500 & 30 & 2.0 & 0.48 & 0.48 & 0.48 & \textbf{0.31} & 0.44 & 0.49 & 0.52 & 0.34 & 0.46\\
A & 1000 & 15 & 0.5 & 0.46 & 0.41 & 0.50 & \textbf{0.36} & 0.48 & 0.48 & 0.53 & 0.55 & 0.61\\
A & 1000 & 15 & 2.0 & 0.32 & 0.32 & 0.28 & \textbf{0.24} & 0.32 & 0.30 & 0.47 & 0.27 & 0.35\\
A & 1000 & 30 & 0.5 & 0.48 & 0.64 & 0.39 & \textbf{0.35} & 0.51 & 0.48 & 0.56 & 0.44 & 0.80\\
A & 1000 & 30 & 2.0 & 0.34 & 0.33 & 0.28 & \textbf{0.26} & 0.29 & 0.28 & 0.46 & \textbf{0.26} & 0.39\\
\midrule
B & 500 & 15 & 0.5 & 2.84 & 2.95 & 2.77 & 2.64 & 3.42 & 2.97 & 4.65 & 2.51 & \textbf{2.48}\\
B & 500 & 15 & 2.0 & 2.53 & 2.45 & 2.20 & 2.43 & 2.37 & 2.34 & 3.07 & 2.23 & \textbf{2.10}\\
B & 500 & 30 & 0.5 & 2.67 & 2.68 & 2.66 & 2.66 & 2.89 & 2.76 & 5.17 & 2.46 & \textbf{2.45}\\
B & 500 & 30 & 2.0 & 2.42 & 2.45 & 2.32 & 2.37 & 2.35 & 2.37 & 3.35 & 2.26 & \textbf{2.12}\\
B & 1000 & 15 & 0.5 & 2.49 & 2.49 & 2.18 & 2.17 & 2.32 & 2.27 & 3.79 & 2.05 & \textbf{1.93}\\
B & 1000 & 15 & 2.0 & 2.30 & 2.25 & \textbf{1.27} & 1.29 & 1.33 & 1.31 & 1.96 & 1.76 & 1.48\\
B & 1000 & 30 & 0.5 & 2.34 & 2.33 & 2.23 & 2.31 & 2.53 & 2.32 & 3.95 & 2.13 & \textbf{1.96}\\
B & 1000 & 30 & 2.0 & 2.31 & 2.33 & 1.50 & 1.67 & 1.58 & 1.62 & 2.17 & 1.92 & \textbf{1.46}\\
\midrule
C & 500 & 15 & 0.5 & 13.48 & 13.42 & 12.90 & 13.52 & 17.62 & 16.92 & 15.02 & 12.50 & \textbf{12.43}\\
C & 500 & 15 & 2.0 & 12.98 & 13.04 & 10.34 & 10.38 & 10.26 & 10.68 & \textbf{9.83} & 11.96 & 11.76\\
C & 500 & 30 & 0.5 & 12.74 & 12.80 & 12.96 & 13.21 & 19.91 & 19.47 & 15.15 & 12.13 & \textbf{12.02}\\
C & 500 & 30 & 2.0 & 12.89 & 12.92 & 11.25 & 11.39 & 12.36 & 12.55 & \textbf{10.70} & 12.08 & 11.87\\
C & 1000 & 15 & 0.5 & 12.81 & 12.81 & \textbf{10.61} & \textbf{10.61} & 13.05 & 13.55 & 10.74 & 11.21 & 10.96\\
C & 1000 & 15 & 2.0 & 12.59 & 12.64 & 6.46 & 6.32 & 6.28 & 6.70 & \textbf{6.26} & 10.16 & 9.79\\
C & 1000 & 30 & 0.5 & 13.09 & 13.10 & 11.70 & 11.69 & 15.66 & 15.84 & 12.35 & 11.93 & \textbf{11.62}\\
C & 1000 & 30 & 2.0 & 12.60 & 12.61 & 7.74 & 7.47 & 7.65 & 7.75 & \textbf{7.27} & 10.96 & 10.07\\
\midrule
D & 500 & 15 & 0.5 & 5.92 & 5.91 & 5.49 & 5.49 & 5.66 & 5.50 & 5.76 & 5.41 & \textbf{5.36}\\
D & 500 & 15 & 2.0 & 5.72 & 5.71 & 4.34 & \textbf{4.22} & 4.25 & 4.30 & 4.31 & 5.22 & 5.14\\
D & 500 & 30 & 0.5 & 5.91 & 6.03 & 5.67 & 5.73 & 5.90 & 5.76 & 6.33 & 5.75 & \textbf{5.65}\\
D & 500 & 30 & 2.0 & 5.63 & 5.54 & \textbf{4.49} & 4.50 & 4.51 & 4.61 & 4.60 & 5.32 & 5.04\\
D & 1000 & 15 & 0.5 & 5.68 & 5.65 & 4.38 & \textbf{4.31} & 4.49 & 4.44 & 4.64 & 4.84 & 4.81\\
D & 1000 & 15 & 2.0 & 5.20 & 5.22 & 2.65 & 2.64 & \textbf{2.53} & 2.68 & 2.74 & 4.22 & 4.03\\

D & 1000 & 30 & 0.5 & 5.80 & 5.70 & \textbf{4.83} & 5.24 & 5.06 & 5.02 & 5.00 & 5.25 & 4.89\\
D & 1000 & 30 & 2.0 & 5.21 & 5.16 & \textbf{2.83} & 2.85 & 2.96 & 2.94 & 2.95 & 4.56 & 4.21\\
\bottomrule
\end{tabular}
\end{table}
\end{landscape}
\restoregeometry

\end{document}